\documentclass[letterpaper]{article}
\usepackage{aaai24}

\usepackage{booktabs} % For formal tables
\usepackage{caption}
\usepackage{multirow}
\usepackage{amsmath}
\usepackage{amsfonts}
\usepackage{times} % DO NOT CHANGE THIS
\usepackage{helvet} % DO NOT CHANGE THIS
\usepackage{courier} % DO NOT CHANGE THIS
\usepackage[hyphens]{url} % DO NOT CHANGE THIS
\usepackage{graphicx} % DO NOT CHANGE THIS
\usepackage{graphicx} % DO NOT CHANGE THIS
\usepackage{natbib} % DO NOT CHANGE THIS
\usepackage{caption} % DO NOT CHANGE THIS
\title{Advancing Ante Hoc Explainable Models through Generative Adversarial Networks}
\author {
    Tanmay Garg,
    Deepika Vemuri,
    Vineeth N. Balasubramanian
}
\affiliations {
    Indian Institute of Technology, Hyderabad, India\\
    gargtanmay1@gmail.com, ai22resch11001@iith.ac.in, vineethnb@cse.iith.ac.in
}
\begin{document}
\maketitle

\begin{abstract}

% Self-Explaining Neural Network (SENN) models have proven adept at unraveling intricate parameter relationships, offering human-readable visualizations. This paper introduces an improved Ante-Hoc method for acquiring concept-based explanations, showcasing enhanced predictive performance and richer insights.

% The proposed model extends SENN by integrating an encoder-generator adversarial framework, akin to Generative Adversarial Networks (GANs), into the architecture, departing from the previous decoder integration attempt. This innovative approach paves the way for more potent and nuanced concept-based learning models. The model appends an explanation generation module to the foundational network, co-training both components. This setup, coupled with the distinctive GAN-based strategy, facilitates unsupervised concept acquisition in an expanded feature space, resulting in heightened classification accuracy and vivid conceptual visualizations for each class.

% This paper elaborates on our framework and presents insights from a diverse array of experiments validating its effectiveness. Furthermore, we conduct an in-depth analysis of how various noise sampling methods influence both classification and concept acquisition.

This paper presents a novel concept learning framework for enhancing model interpretability and performance in visual classification tasks. Our approach appends an unsupervised explanation generator to the primary classifier network and makes use of adversarial training. During training, the explanation module is optimized to extract visual concepts from the classifier's latent representations, while the GAN-based module aims to discriminate images generated from concepts with true images. This joint training scheme enables the model to implicitly align its internally learned concepts with human-interpretable visual properties. Comprehensive experiments demonstrate the robustness of our approach, while producing coherent concept activations. We analyse the learned concepts, showing their semantic concordance with object parts and visual attributes. We also study how perturbations in the adversarial training protocol impact both classification and concept acquisition.
In summary, this work presents a significant step towards building inherently interpretable deep vision models with task-aligned concept representations - a key enabler for developing trustworthy AI for real-world perception tasks.
% To facilitate interpretation and validation of the model, we develop a novel visualization interface for exploring image-specific concept attribution maps. Extensive user studies show the concepts help humans predict classifications, indicating they reflect meaningful visual explanations intrinsic to the network.

\end{abstract}

% \keywords{Generative Adversarial Networks, Machine Learning, Ante-Hoc, Explainable Models}

\maketitle
\section{Introduction}\label{sec:intro}
% Deep Neural Network (DNN) models have revolutionized fields such as Computer Vision \cite{VGG}, Natural Language Processing \cite{gpt3}, Healthcare \cite{health}, and Finance \cite{finance}. These models have advanced to great extents at handling complex tasks such as image recognition, machine translation, and anomaly detection.

% However, these DNNs are inherently black-boxes; with the increasing complexity of these models leading to a lack of transparency and interpretability. This black-box nature of DNNs has been of great concern in the scientific community, especially in risk-sensitive scenarios such as healthcare or criminal justice. In the case of healthcare, for instance, many patients would want to know why a disease diagnosing model gives a certain output. False negatives and positives would also be easier to diagnose and validate, which could have potentially severe consequences if undetected.

Deep neural networks (DNNs) have ushered in a revolution across domains like Computer Vision \cite{VGG}, Natural Language Processing \cite{gpt3}, Healthcare \cite{health}, and Finance \cite{finance}. They have made significant strides in handling intricate tasks like image recognition, machine translation, and anomaly detection. However, they come with a challenge - they are essentially black-box systems. The increasing complexity of these models has led to a lack of transparency and interpretability \cite{lipton, ravikumar}. This opacity has raised significant concerns within the scientific community, particularly in critical areas like healthcare and criminal justice. In healthcare, for instance, patients would want to know why a disease-diagnosing model provided them with a certain result.
%In healthcare, for instance, it's crucial for patients to understand why a disease-diagnosing model provides a specific result. 
Additionally, being able to identify and verify false positives and negatives is essential, as such oversight could have potentially serious consequences.

Explainable models have become instrumental in establishing transparency, a key factor in building trust with users. 
%Their emergence has brought about a fresh perspective to the explainability of Deep Neural Networks (DNNs). 
In recent years, there has been a surge of research in this area, with most works coming under two broad categories: post-hoc and ante-hoc methods.

\textit{Post-hoc explainability} methods attempt to provide explanations as a separate module on already trained models.
Saliency maps \cite{saliency_maps} are a prime example of this line of work, introducing a method that visually highlights the points on an image that activate neurons, depending on the predicted class by the deep neural network.  However, decoupling the explanation method from the explained model makes it a challenge to discern whether the model's prediction was incorrect or the explanation provided was at fault.

\textit{Ante-hoc explainability} methods, on the other hand, provide explanations implicitly during model training itself. There have been several ante-hoc works in recent times that make use of \textit{concepts} \cite{CBM, efros, efros2}. These methods assume that each class can be broken down into a set of concepts, i.e. that concepts can be used to signify the distinctive features or characteristics that make up a particular class. For example, in the case of MNIST \cite{MNSIT}, concepts could include straight lines, types of curves in the digits, or even more specific patterns that may appear in a digit. 
Self-explaining neural networks (SENN) \cite{SENN} exemplify such approaches, offering a straightforward means to acquire interpretable concepts by extending a linear predictor. When presented with an input image, the prediction is generated based on a weighted combination of these concepts. 
\cite{Sarkar2021AFF}, introduce a method to account for varying degrees of concept supervision in a SENN-like framework.

In this paper, we build upon and extend the findings of \cite{Sarkar2021AFF}, showing how introducing an adversarial component into the framework can better guide representation learning. We propose a modified loss, harness the benefits of randomization and use labels as supplementary information for conditioning the reconstruction process.
%and introduce an improved method for concept acquisition by leveraging Generative Adversarial Networks (GANs) \cite{GAN}. With GANs, we propose a modified loss function. 
%One of the limitations observed in \cite{Sarkar2021AFF} pertained to its approach in image reconstruction, utilizing a basic decoder that neither harnessed the benefits of randomization nor utilized labels as supplementary information for conditioning the reconstruction process. Our work endeavors to address these limitations.

\begin{figure*}[h!]
    \centering
\includegraphics[width=1.8\columnwidth]{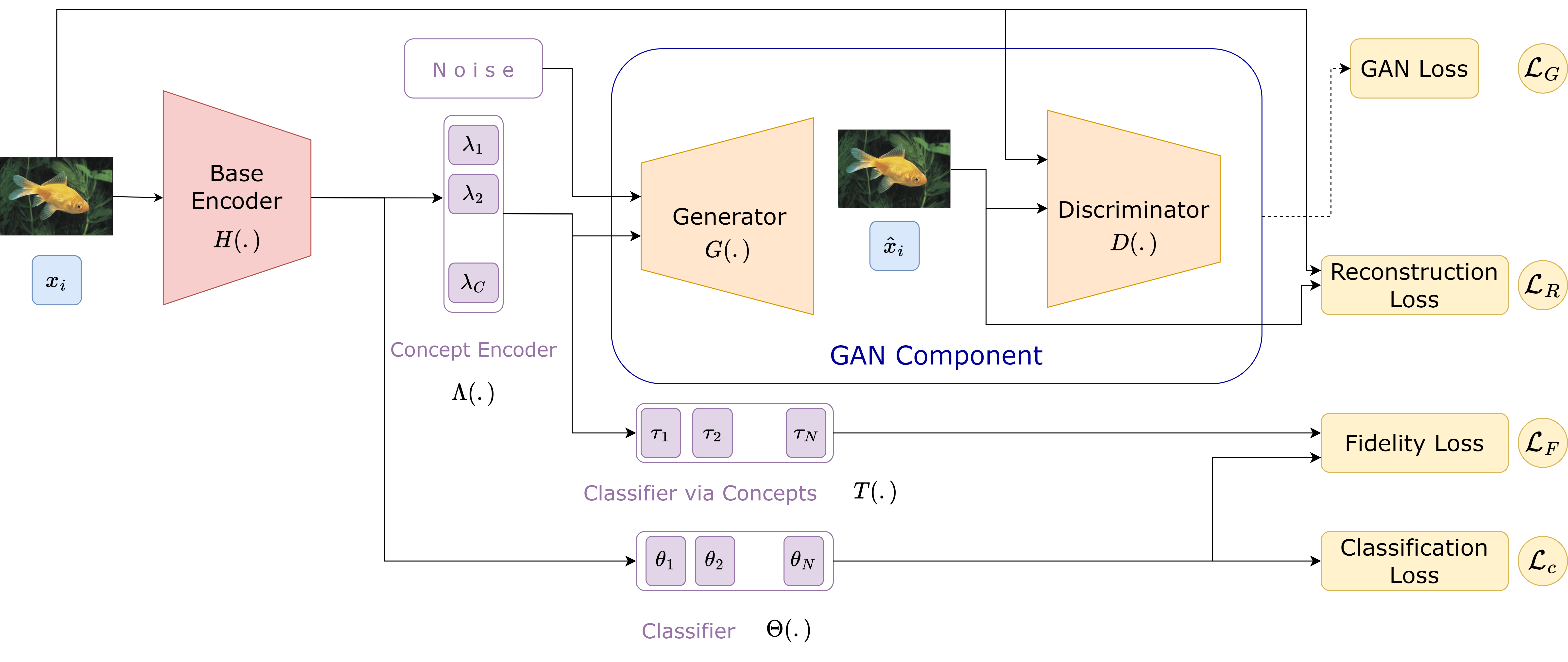}
    \caption{Overview of our Proposed Architecture. N is the number of classes, C is the number of concepts}
    \label{fig:senn_gan}
\end{figure*}

% To summarize, we have made the following contributions:
% \begin{itemize}
%     \item Introduced a new and modified architecture that shows an improvement in performance in comparison to \cite{Sarkar2021AFF}. The crux of our novelty lies in  utilizing a Generated Adversarial Network \cite{GAN} within the architecture
%     \item We have conducted multiple experiments to study and compare how different GANs such as Vanilla GAN \cite{GAN} and cGAN \cite{CGAN} can affect the performance and concept visualization
%     \item We have experimented with different noise generation methods to understand how noise sampled from a gaussian distribution can affect the generation of concepts for our framework
%     \item Our method leverages the adversarial nature of GANs and our noise method to generate better images that could enforce better concept encoding
%     \item The discriminator of GAN helps in maintaining the similarity between the input image and the reconstructed image
% \end{itemize}
To summarize, our contributions are as follows:
\begin{itemize}
    \item We introduce a novel, enhanced architecture that demonstrates improved performance compared to the baselines. The key aspect here lies in the integration of a Generative Adversarial Network (GAN) \cite{GAN} within the architecture.
    \item We conduct a series of experiments to analyze and compare the impact of different GAN variants, such as a vanilla GAN \cite{GAN} and conditional GAN (cGAN) \cite{CGAN}, on performance and concept visualization.
    \item We explore various methods for generating noise to understand how noise sampled from a Gaussian distribution influences concept generation in our framework.
    \item Our approach capitalizes on the adversarial nature of GANs and noise generation method to produce higher-quality images that facilitate more robust concept encoding.
    % \item The discriminator component of the GAN aids in maintaining similarity between the input image and the reconstructed image.
\end{itemize}

\section{Related Work}\label{sec:rel_work}

% Our objective was to develop an improved method that can generate effective concepts to represent the classes. 

% \textbf{Post-Hoc Methods:} In the case of Saliency Maps \cite{saliency_maps}, the gradients were computed with respect to the final predictions of the input images. This gradient information facilitates the generation of a heatmap, which, when overlaid onto the input image, highlights the pixels that have activated the neurons. This technique represents a Post-Hoc methodology for DNN interpretation, focusing on explanations after model predictions are made. However, there are multiple limitations to their approach, one of which is its susceptibility to minor input pixel perturbations, that is capable of exerting influence on the resultant saliency map. 

% In the case of Grad-CAM \cite{GradCAM}, a \textit{class-discriminative localization technique} was introduced that would interpret CNN-based DNN models without altering the model architecture. It utilizes the gradient information to highlight regions contributing significantly to class prediction. However, Grad-CAM encounters challenges in localizing multiple instances of an object within an image and exhibits imprecise heatmap localization for class region coverage. This is due to the inherent reliance on partial derivatives. Moreover, the iterative upsampling and downsampling operations within Grad-CAM's process might lead to a substantial loss of information.

%Our primary objective was to devise an enhanced method capable of generating meaningful concepts to represent distinct classes.

\textbf{Post-Hoc Methods:} In addition to Saliency Maps \cite{saliency_maps} and Grad-CAM \cite{GradCAM}, other influential post-hoc techniques include LIME \cite{guestrin} and DeepLift \cite{deeplift}. LIME provides model-agnostic local explanations by approximating any classifier with an interpretable linear model. DeepLift decomposes the predictions of a Deep Neural Network by backpropagating contribution scores to the inputs. Most of these methods are gradient-based, with the problem of the root cause  of errors being difficult to diagnose.

\textbf{Concept-based Models: }\cite{CBM} propose concept bottleneck models for deep learning interpretability, amenable to test-time human intervention. Concept bottleneck models have an intermediate concept layer and are trained on human-specified interpretable concepts in addition to the task labels. 
They undergo a two-step training process where the inputs predict concepts and the concepts predict the label.
%It systematically studies different variants of concept bottleneck models, comparing them against standard end-to-end models. They explore training approaches like independent, sequential and joint training of the concept and task prediction parts. The paper evaluates the models on medical and fine-grained image classification tasks, demonstrating comparable task accuracy while achieving high concept accuracy. 
The interpretable nature of concept bottleneck models enables test-time interventions, where human experts can correct the concept values of a wrong prediction to simulate rich model-human collaboration. The effectiveness of such interventions varies based on the training approach, highlighting the need to study factors beyond just task/concept accuracy. %Their work establishes concept bottleneck models as a promising approach for building interpretable models tailored for human-AI collaboration.
There have been several concept-based models like \cite{interactive, zarlenga2022concept, yuksekgonul2022posthoc}. However, there has been some criticism as to whether these models really learn as intended \cite{margeloiu2021concept}.

\textbf{Prototype-based Learning: }\cite{prototype_paper} introduce an approach for interpreting deep neural networks by integrating an autoencoder with a \textit{prototype} layer during training. The model classifies inputs based on their proximity to encoded examples in the prototype layer, facilitating an intuitive case-based reasoning mechanism. The jointly optimized prototypes, guided by various loss terms, connect the network's decisions with explanations, visible through visualization of class-representative prototypes. While achieving competitive accuracy compared to CNN baselines, the method offers integrated explanations without post-hoc techniques. 
%This work pioneers interpretable deep learning by embedding explanations into model design and training, laying the foundation for ongoing advancements in explainable deep neural networks.
Several extensions to prototype-based methods have been proposed, like \cite{prototype_1, Donnelly_2022_CVPR}

\textbf{Other methods: } Several prior studies have developed methods focusing on both high accuracy and explainability. Some methods take the approach of making use of auto-encoders that could enable the reconstruction of images such as \cite{efros2}. There are human-in-the-loop works related to involving human feedback and developing concepts that both align with human's intuition of concept such as \cite{concept_human}. Several derivatives of prototype-based models have proven to be quite impressive at visualizing representations and explanations \cite{prototype}.

\subsection{Background}\label{sec:background}

\textbf{Self Explaining Neural Networks \cite{SENN}:} This was one of the first works to propose a robust ante-hoc framework along with metrics to measure interpretability. Starting with a simple linear regression model, which is inherently interpretable - given that the model's parameters are linearly related, the paper successively generalizes it to more complicated models like neural networks.
%For image tasks, additional factors like color, hue, saturation, etc., need to be considered for explainability. Therefore, for image classification, they use neural networks instead of simple linear regression models. 
With neural networks, several considerations come into play:

\begin{itemize}
    \item The concepts representing an image should retain the information the image holds.
    \item The concepts visualized for classification should be distinct from one another.
    \item The learned and visualized concepts should be human-understandable.
\end{itemize}

To address these points, they employ an auto-encoder to encode the input image into relevant concepts, while using a combination of reconstruction loss and classification loss to optimize the model. One limitation we observe is that the approach derives concepts solely based on the dataset without utilizing additional information like labels or extra images. Extensions to this work have attempted to enhance concept explainability by disentangling and contrastively learning them \cite{CSENN}. Our work, on the other hand, leverages generative models to create more refined concepts. \\
 
% \\

% \textbf{Ante-Hoc Explainable via Concepts:} In paper \cite{Sarkar2021AFF}, a method was introduced that could visualize concepts in an Ante-Hoc unsupervised manner, i.e. concepts and interpretability could be obtained during the training process itself. The proposed framework could be  appended to an existing backbone model architecture and be jointly trained. The framework introduces the ability to incorporate differing degrees of of supervision such as fully supervised and unsupervised concept learning. They were able to achieve this by adding very few additional parameters, and some extra terms in the loss calculation that helped the model train in an Ante-Hoc fashion. One limitation in this work was that they were attempting to use a simple encoder-decoder architecture that does not take the similarity of the decoded images with respect to the input image into consideration. Moreover, they do not consider label information to train the encoder to better encode the images.

\textbf{Ante-Hoc Explainability via Concepts \cite{Sarkar2021AFF}:} In this work, an ante-hoc framework, allowing for different levels of supervision, including fully supervised and unsupervised concept learning, was proposed, building upon \cite{SENN}. 
%This means that concepts and explainability could be understood during the training process itself. 
The framework could be added to any existing backbone model and optimized jointly. The paper primarily introduces the notion of \textit{fidelity loss} as a way to capture semantics into concepts and a way to visualize the concepts learn by the model. 
In our work, we assume the basic setup from this paper and make specific modifications to enhance it.

\section{Methodology}\label{sec:method}

\begin{table*}[ht!]
\centering
\scalebox{1}{
\begin{tabular}{c|c|c|c|c}
\toprule
Model                    & VGG Model & Method & Accuracy       & Auxiliary Accuracy \\ \midrule
Baseline                & NA        & NA     & 64.46          & 44.65              \\
SENN                     & NA        & NA     & 36.57          & NA                 \\ \midrule \midrule
 Vanilla GAN (B=32, S=10) & 11        & DAN    & \textbf{65.49} & {45.05}     \\ 
% Vanilla GAN (B=32, S=10) & 11        & ICN    & \textbf{65.13} & \textbf{46.16}     \\ 
% Vanilla GAN (B=32, S=10) & 11        & PCN    & \textbf{65.47} & \textbf{45.80}     \\ 
Vanilla GAN (B=32, S=10) & 19        & DAN    & 60.71          & 44.52              \\
cGAN (B=32, S=10)        & 11        & DAN    & {65.37} & \textbf{45.36}     \\ 
% cGAN (B=32, S=10)        & 11        & ICN    & \textbf{65.15} & 41.69              \\ 
% cGAN (B=32, S=10)        & 11        & PCN    & \textbf{65.21} & \textbf{45.44}     \\
cGAN (B=32, S=10)        & 19        & DAN    & {64.42} & {44.92}     \\ 
% cGAN (B=32, S=10)        & 19        & ICN    & 62.42 &	\textbf{44.87}              \\ 
% cGAN (B=32, S=10)        & 19        & PCN    & 64.00 &	44.56     \\ \midrule
 \bottomrule
% Vanilla GAN (B=32, S=10) & 19        & ICN    & 58.09          & 43.78              \\ 
% Vanilla GAN (B=32, S=10) & 19        & PCN    & 60.02          & \textbf{46.19}     \\ \bottomrule
\end{tabular}
}
\caption{\textit{Accuracy} (in \%) and \textit{Auxiliary Accuracy} (in \%) for comparison with the baseline and SENN on CIFAR100. cGAN = Conditional GAN, B = Batch Size, S = size of noise. Our method classifies better.}
\label{tab:cifar100_all}
\end{table*}

% \begin{table*}[]
% \scalebox{0.9}{
% \begin{tabular}{c|c|cc|cc|cc}
% \toprule
% \multirow{2}{*}{Model}   & \multirow{2}{*}{VGG Model} & \multicolumn{2}{c}{DAN} & \multicolumn{2}{c}{ICN} & \multicolumn{2}{c}{PCN} \\ 
%  &
%    &
%   \multicolumn{1}{c}{Accuracy} &
%   \multicolumn{1}{c|}{Aux. Accuracy} &
%   \multicolumn{1}{c}{Accuracy} &
%   \multicolumn{1}{c|}{Aux. Accuracy} &
%   \multicolumn{1}{c}{Accuracy} &
%   \multicolumn{1}{c}{Aux. Accuracy} \\ \midrule
% \multicolumn{1}{c|}{cGAN (B=32, S=10)} &
%   \multicolumn{1}{c|}{11} &
%   \multicolumn{1}{c}{\textbf{65.37}} &
%   \multicolumn{1}{c|}{45.36} &
%   \multicolumn{1}{c}{65.15} &
%   \multicolumn{1}{c|}{41.69} &
%   \multicolumn{1}{c}{65.21} &
%   \multicolumn{1}{c}{{45.44}} \\ 
% \multicolumn{1}{c|}{cGAN (B=32, S=10)} &
%   \multicolumn{1}{c|}{19} &
%   \multicolumn{1}{c}{\textbf{64.42}} &
%   \multicolumn{1}{c|}{{44.92}} &
%   \multicolumn{1}{c}{62.42} &
%   \multicolumn{1}{c|}{44.87} &
%   \multicolumn{1}{c}{64.00} &
%   \multicolumn{1}{c}{44.56} \\ 
% \multicolumn{1}{c|}{Vanilla GAN (B=32, S=10)} &
%   \multicolumn{1}{c|}{11} &
%   \multicolumn{1}{c}{\textbf{65.49}} &
%   \multicolumn{1}{c|}{45.05} &
%   \multicolumn{1}{c}{65.13} &
%   \multicolumn{1}{c|}{46.16} &
%   \multicolumn{1}{c}{65.47} &
%   \multicolumn{1}{c}{{45.80}} \\ 
% Vanilla GAN (B=32, S=10) & 19                         & \textbf{60.71}      & 44.52      & 58.09      & 43.78      & 60.02      & {46.19}      \\ \bottomrule
% \end{tabular}
% }
\begin{table*}[]
\scalebox{0.9}{
\begin{tabular}{c|c|cc|cc|cc}
\toprule
\multirow{2}{*}{Model}   & \multirow{2}{*}{VGG Model} & \multicolumn{2}{c}{DAN} & \multicolumn{2}{c}{ICN} & \multicolumn{2}{c}{PCN} \\ 
 &
   &
  \multicolumn{1}{c}{Accuracy} &
  \multicolumn{1}{c|}{Aux. Accuracy} &
  \multicolumn{1}{c}{Accuracy} &
  \multicolumn{1}{c|}{Aux. Accuracy} &
  \multicolumn{1}{c}{Accuracy} &
  \multicolumn{1}{c}{Aux. Accuracy} \\ \midrule
  \multicolumn{1}{c|}{Vanilla GAN (B=32, S=10)} &
  \multicolumn{1}{c|}{11} &
  \multicolumn{1}{c}{\textbf{65.49}} &
  \multicolumn{1}{c|}{45.05} &
  \multicolumn{1}{c}{65.13} &
  \multicolumn{1}{c|}{46.16} &
  \multicolumn{1}{c}{65.47} &
  \multicolumn{1}{c}{{45.80}} \\
\multicolumn{1}{c|}{Vanilla GAN (B=32, S=10)} &
  \multicolumn{1}{c|}{19} &
  \multicolumn{1}{c}{\textbf{60.71}} &
  \multicolumn{1}{c|}{44.52} &
  \multicolumn{1}{c}{58.09} &
  \multicolumn{1}{c|}{43.78} &
  \multicolumn{1}{c}{60.02} &
  \multicolumn{1}{c}{{46.19}} \\ 
\multicolumn{1}{c|}{cGAN (B=32, S=10)} &
  \multicolumn{1}{c|}{11} &
  \multicolumn{1}{c}{\textbf{65.37}} &
  \multicolumn{1}{c|}{45.36} &
  \multicolumn{1}{c}{65.15} &
  \multicolumn{1}{c|}{41.69} &
  \multicolumn{1}{c}{65.21} &
  \multicolumn{1}{c}{{45.44}} \\ 
\multicolumn{1}{c|}{cGAN (B=32, S=10)} &
  \multicolumn{1}{c|}{19} &
  \multicolumn{1}{c}{\textbf{64.42}} &
  \multicolumn{1}{c|}{{44.92}} &
  \multicolumn{1}{c}{62.42} &
  \multicolumn{1}{c|}{44.87} &
  \multicolumn{1}{c}{64.00} &
  \multicolumn{1}{c}{44.56} \\ 
\bottomrule
% Vanilla GAN (B=32, S=10) & 19                         & \textbf{60.71}      & 44.52      & 58.09      & 43.78      & 60.02      & {46.19}      \\ \bottomrule
\end{tabular}
}
\caption{\textit{Accuracy} (in \%) and \textit{Auxiliary Accuracy} (in \%) for comparing our models on CIFAR100. cGAN = Conditional GAN, B = Batch Size, S = size of noise. Aux. Accuracy = Auxiliary Accuracy.}
\label{tab:cifar100_compare}
\end{table*}

\subsection{Proposed Architecture}\label{sec:arch}
A typical deep learning classification pipeline has a base encoder function followed by a classifier function. Let $\mathcal{X}$ be the input space and $\mathcal{Y}$ be the label space for our training set $\mathcal{D} = \{x_i, y_i\}_{i=1}^N$, sampled i.i.d from some source distribution $\mathcal{P} : \mathcal{X} \times\mathcal{Y}$, where $\mathcal{X} \in \mathrm{R}^d$, and $\mathcal{Y}$ is a one-hot encoded vector. The base encoder $H(.)$, extracts representation vectors that are then fed into the classifier $\Theta(.)$. One way of making such a setup interpretable is to provide explanations via concepts. In order to do this, a concept encoder $\Lambda(.)$ is introduced in the pipeline. $\Lambda(.)$ takes in the representations extracted by $H(.)$ and learns a set of concepts $\{\lambda_1, \lambda_2,…\lambda_C\}$ that are used to explain the classification by passing them through $T(.)$. The concepts are \textit{latents} that represent the attributes of a class. In our framework, we consider the latents to be scalars representing the degree to which a concept is present in a given image, i.e they are a score. The predictions given by $T$ and $\Theta$ should match, which is enforced through a fidelity loss $\mathcal L_F$. Further, the concepts learnt are passed through a decoder to reconstruct the input image and as such are enforced to capture image semantics.

Building upon this setup, we have developed a novel architecture incorporating a Generative Adversarial Network (GAN) \cite{GAN} into the framework. GANs consist of two core components: a generator and a discriminator. The primary task of the generator is to fabricate synthetic images or representations that closely mimic a specific dataset or probability distribution. While the discriminator, on the other hand, is responsible for discerning whether an image is authentic or a generated clone. In the above pipeline, we propose sending the concepts to a GAN $(G(.) , D(.))$, making use of the adversarial mechanism to retrieve better concepts. $G(.)$ takes in the concepts, supplemented by some noise. The noise introduces a degree of randomness (discussed in Section 4.1), which along with the concepts is used to generate an image using deconvolution operations. $\mathcal L_R$ is needed to enforce that the concepts capture semantics (similar to the role of the decoder in \cite{Sarkar2021AFF}, with noise as an additional input). $D(.)$ takes in an image and outputs whether it is real or fake. So, the loss now becomes:

\begin{equation}
\begin{aligned}\label{eq:our_loss}
    \mathcal{L} = & \, \mathcal{L}_c + \mathcal{L}_R
    + \mathcal{L}_F + \mathcal{L}_G
\end{aligned}
\end{equation}

where, $\mathcal{L}_c$ is the classification loss, $\mathcal{L}_R$ is the reconstruction loss between the input image and the generated image, $\mathcal{L}_F$ is the fidelity loss and $\mathcal L_G$ is the GAN loss.

The discriminator $D(.)$ takes a real image $x_i$ and performs a forward pass. The loss and the gradients are then calculated. $D(.)$ also performs a forward pass on the generated image $\hat{x}_i$, after which the loss and gradients are calculated. The gradients calculated are passed through $D(.)$ and $G(.)$ as they aren't detached before being sent to the discriminator. This interconnection ensures that both the generator and discriminator are jointly optimized, working together to produce more convincing fake images while still accurately detecting them. This finishes a single iteration of training the network and the parameters of generator and discriminator are updated for the next round of training.

The overall loss function that we would optimize is:

\begin{equation}
\begin{aligned}\label{eq:loss}
    \mathcal{L} = & \, \alpha\mathcal{L}_c\left(y_i, \hat{y}_i\right) + \beta \mathcal{L}_R\left(x_i, \hat{x}_i\right)\\
    &+ \gamma \mathcal{L}_F\Bigl(\Theta\bigl(H(x_i)\bigr), T\bigl(\Lambda\bigl(H(x_i)\bigr)\bigr)\Bigr) \\
    & + \delta\mathbb{E}_{x_i}\left[\log\,\Bigl(D\left(x_i\right)\Bigr)\right]\\ &+ \delta\mathbb{E}_{n}\left[\log\,\Bigl(1-D\bigl(G\left(\{\Lambda\left(H(x_i)\right), n\}\right)\bigr)\Bigr)\right]
\end{aligned}
\end{equation}\\

In the Eq \ref{eq:loss}, $\mathcal{L}_c$ is cross entropy loss, $\mathcal{L}_R$ is the $L_2$ loss between the input image and the generated image, $\mathcal{L}_F$ is the MSE between the outputs of $\Theta$ and $T$ \cite{SENN, Sarkar2021AFF, bastani}, and the remaining two terms are the GAN loss $\mathcal{L}_G$. $\alpha, \beta, \gamma, \delta$ are arbitrary terms for weighting, that have been introduced from an implementation and empirical perspective. 
% $\alpha$ and $\beta$ are arbitrary terms for linear combination that have been introduced from an implementation and empirical perspective, and 
$n$ is the noise that has been sampled using methods in Section \ref{sec:data_noise}.

The rationale behind using GANs stems from the desire to provide a substantially broader spectrum for the concepts to be learnt from. In a GAN framework, the input to the generator is typically sampled from a normal distribution. However, in our methodology, we adopt an approach that combines inputs from both a normal distribution and the encoding. This choice effectively enlarges the feature space available to the generator. Consequently, it empowers the generator to adeptly reconstruct images from the encoding. This, we posit, ultimately leads to an enhancement in the encoder's proficiency in generating more informative encodings.

\begin{table*}[ht!]
\centering
\begin{tabular}{c|c|c|c|c}
\toprule
Model                    & VGG Model & Method & Accuracy & Auxiliary Accuracy \\ \midrule
Baseline                 & NA        & NA     & 91.68    & 90.86              \\ 
SENN                     & NA        & NA     & 84.50     & 84.50               \\ \midrule \midrule
Vanilla GAN (B=32, S=10) & 8         & ICN    & 91.57    & 89.63              \\ 
Vanilla GAN (B=32, S=10) & 11        & ICN     & 91.66    & 90.04              \\ 
cGAN(B=32, S=10)         & 8         & PCN     & 91.60     & 89.94              \\ 
cGAN(B=32, S=10)         & 11        & DAN     & 91.58    & 89.90               \\ 
cGAN(B=32, S=5)          & 19        & DAN     & \textbf{91.82}    & 90.23              \\ \bottomrule
\end{tabular}
\caption{\textit{Accuracy} (in \%) and \textit{Auxiliary Accuracy} (in \%) for comparison with the baseline and SENN on CIFAR10. cGAN = Conditional GAN, B = Batch Size, S = size of noise. Our method classifies better.}
\label{tab:res}
\end{table*}

\begin{table*}[ht!]
\centering
\scalebox{0.91}{
\begin{tabular}{c|c|cc|cc|cc}
\toprule
\multirow{2}{*}{Model} &
  \multirow{2}{*}{VGG Model} &
  \multicolumn{2}{c}{DAN} &
  \multicolumn{2}{c}{ICN} &
  \multicolumn{2}{c}{PCN} \\
 &
   &
  \multicolumn{1}{c}{Accuracy} &
  Aux. Accuracy &
  \multicolumn{1}{c}{Accuracy} &
  Aux. Accuracy &
  \multicolumn{1}{c}{Accuracy} &
  Aux. Accuracy \\ \midrule
Vanilla GAN  (B=32, S=10) & 8  & \multicolumn{1}{c}{91.53} & 90.00    & \multicolumn{1}{c}{\textbf{91.57}} & 89.63 & \multicolumn{1}{c}{91.41} & 89.57 \\
Vanilla GAN  (B=32, S=10) & 11 & \multicolumn{1}{c}{91.38} & 89.63 & \multicolumn{1}{c}{90.80}  & 89.37 & \multicolumn{1}{c}{\textbf{91.66}} & 90.94 \\ 
cGAN  (B=32, S=10)        & 8  & \multicolumn{1}{c}{91.55} & 90.15 & \multicolumn{1}{c}{91.44} & 90.13 & \multicolumn{1}{c}{\textbf{91.60}}  & 89.94 \\ 
cGAN (B=32, S=10)         & 11 & \multicolumn{1}{c}{\textbf{91.58}} & 89.90  & \multicolumn{1}{c}{91.34} & 89.96 & \multicolumn{1}{c}{91.28} & 89.82 \\ 
cGAN (B=32, S=5)          & 11 & \multicolumn{1}{c}{91.36} & 89.74 & \multicolumn{1}{c}{91.44} & 89.77 & \multicolumn{1}{c}{\textbf{91.47}} & 89.68 \\ 
cGAN (B=32, S=10)         & 19 & \multicolumn{1}{c}{91.52} & 90.09 & \multicolumn{1}{c}{\textbf{91.76}} & 90.08 & \multicolumn{1}{c}{91.45} & 89.99 \\ 
cGAN (B=32, S=5)          & 19 & \multicolumn{1}{c}{\textbf{91.82}} & 90.23 & \multicolumn{1}{c}{91.15} & 89.62 & \multicolumn{1}{c}{91.44} & 89.60  \\ \bottomrule
\end{tabular}}
\caption{\textit{Accuracy} (in \%) and \textit{Auxiliary Accuracy} (in \%) for comparing our models on CIFAR10. cGAN = Conditional GAN, B = Batch Size, S = size of Noise. Aux. Accuracy = Auxiliary Accuracy.}
\label{tab:all_model}
\end{table*}

\section{Experiments and Results}\label{sec:exp}

% The paper on Ante-Hoc Explainability \cite{Sarkar2021AFF} serves as our baseline, which surpasses SENN \cite{SENN} on CIFAR-10 \cite{CIFAR10}.
% %has kept their baselines as SENN\cite{SENN} for CIFAR-10\cite{CIFAR10} dataset in an unsupervised learning environment, and 
% Since our results outperform \cite{Sarkar2021AFF}, we have taken the liberty to set our baseline as \cite{Sarkar2021AFF} itself rather than SENN \cite{SENN} - in a similar scenario in an unsupervised learning environment. However, we have included the results of SENN as well for comparison.

We carry out a set of experiments to compare and demonstrate the performance of our framework. We show several variations of models. The baseline for our work is the ante-hoc explainability method proposed by \cite{Sarkar2021AFF}, which builds upon and outperforms SENN \cite{SENN} on the CIFAR-10 and CIFAR-100 datasets. We reimplement their method using their code and hyperparameters, while taking care to control factors like network architecture, training procedure, and hardware to ensure differences are solely due to the methodology.
%In our experiments, we choose to set our baseline as \cite{Sarkar2021AFF} due to its superior performance in a similar unsupervised learning environment. 
However, we also include results from SENN for comparative analysis. For the SENN results, we pick the results reported in \cite{Sarkar2021AFF}\textit{(They do not report aux. acc. on CIFAR100)}.
%In this section, we conduct a comprehensive comparison between our best models with the our baseline \cite{Sarkar2021AFF}. 
Our evaluation criteria includes $\Theta$ classification accuracy (top 1\% accuracy) and $T$ classification accuracy,
%the maintenance of concept faithfulness, 
which we refer to as accuracy and auxiliary accuracy respectively. Given the necessity of maintaining faithful and explainable concepts, while providing accurate image classification, we give higher importance to prediction accuracy, followed next by auxiliary accuracy. 
We briefly describe our evaluation criteria below:
%Since the task of image classification is the primary objective without losing out on the faithfulness and the explainability of the concepts, we give more weightage to the accuracy of the prediction followed by the auxiliary accuracy. We briefly describe our evaluation criteria below:

\textbf{Accuracy} (\textit{top 1\% accuracy})\textbf{:} This metric corresponds to the classification accuracy for the input images with respect to the ground truth labels.

\textbf{Auxiliary Accuracy \cite{SENN}:} This metric corresponds to the classification accuracy %meaningfulness of the acquired concepts. This is evaluated 
based on the output from $T(.)$ and its proficiency in predicting the class labels and indirectly measures the meaningfulness of the acquired concepts.

% \begin{figure*}
%     \centering
%     \includegraphics[width=1.8\columnwidth]{images/cgan19.jpg}
%     \caption{Top 5 images that activate the learnt concepts using cGAN (VGG 19) DAN (Method 1) (B=32, S=5) on CIFAR10. Eg: Cpt 9 captures a concept corresponding to the grey color, Cpt 10 corresponds to frog skin.}
%     \label{fig:cgan19}
% \end{figure*}

\begin{figure*}
    \centering
    \includegraphics[width=1.8\columnwidth]{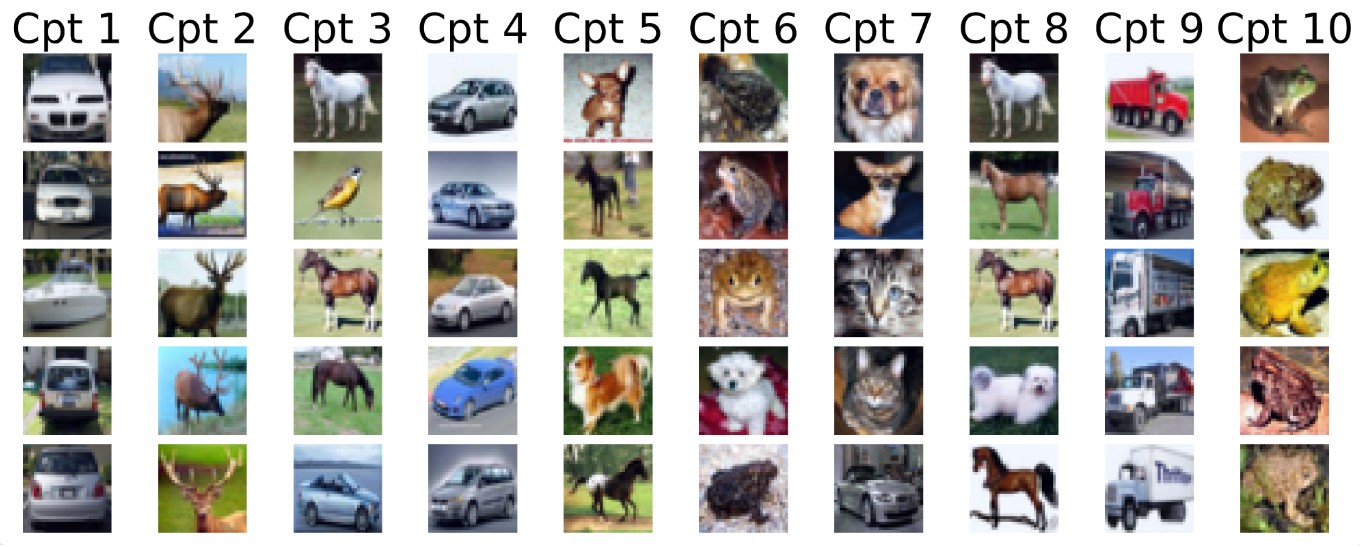}
    \caption{Top 5 images for CIFAR10 that activate the learnt concepts using cGAN (VGG 11) DAN  (B=32, S=10). Eg: Cpt 2 captures antlers, Cpt 1 captures the color white - here we see that activated images are from different classes (ship, car).}
    \label{fig:cgan11}
\end{figure*}

\subsection{Datasets and Comparison Methods}\label{sec:data_noise}

\textbf{Datasets}: For our experiments, we choose the CIFAR-10 and CIFAR-100 \cite{CIFAR10} benchmarks to facilitate comparisons with prior work. CIFAR-10 consists of 60,000 32x32 coloured images from 10 classes, with 6,000 images per class. The dataset split of 50,000/10,000 train/test images providing sufficient data for training deep networks while maintaining a separate test set for unbiased evaluation. CIFAR-100 is slightly more challenging, containing the same number of images but partitioning them into 100 classes, each with 600 images. This increased class variability and lower samples per class simulate real-world fine-grained classification challenges more closely. Compared to CIFAR-10, CIFAR-100 tests a model's ability to discriminate between subtle inter-class differences.

% The CIFAR-10 \cite{CIFAR10} dataset consists of 60000 32x32 colour images in 10 classes, with 6000 images per class. The entire dataset is divided into 50000 training images and 10000 test images.
We chose CIFAR-10 and CIFAR-100 as their moderate sizes allowed us to conduct extensive experiments in reasonable time to thoroughly test different architectures and design decisions, as compared to large-scale datasets such as ImageNet \cite{IMAGENET}. Both the datasets contain complex and diverse real life objects in a variety of backgrounds. 
% The CIFAR-10 dataset was chosen because of its real world complexity and diverse real life objects in various backgrounds. The dataset is relatively small as compared to huge dataset such as ImageNet \cite{IMAGENET} but has been used for benchmarking classification models. This smaller scale offers the advantage of conducting experiments with multiple models and architectures in a time-efficient manner.

\textbf{Noise Methods}: This study introduces a framework with an emphasis on the integration of noise into the network, typically drawn from a Normal distribution $\mathcal{N}(0,1)$. The impact of various noise sampling techniques on GAN training has been thoroughly studied. Given the batch size $B$ of images as 32, concepts of size 10, and a designated noise size $S$ of 5, the shape of the concepts and the noise would be \texttt{32x10x1} (\(B \times C \times 1\)) and \texttt{32x5x1} (\(B \times S \times 1\)), respectively. Once the noise is added, the final shape of the concepts would be \texttt{32x15x1} (\(B \times (C+S) \times 1\)). Different noise generation strategies are explored. The noise sampled helped us in determining the effects of different noise perturbations on our model.

\textbf{Method 1: Direct Align Noise (DAN):} The noise is directly aligned with batch size and noise length. Consequently, the sampled noise follows the shape (\(B \times S \times 1\)) where the entire (\(B \times S \times 1\)) matrix is sampled from $\mathcal{N}(0,1)$ and together represents a Gaussian.

\textbf{Method 2: Iterative Concat Noise (ICN):} The noise is sampled multiple times to achieve the desired dimension. Initially, noise is sampled as \(B \times 1 \times 1\) and then concatenated \(S\) times. Each row of size \(B \times 1 \times 1\) is sampled from a Gaussian.

\textbf{Method 3: Progressive Concat Noise (PCN):} The noise is sampled multiple times, but it begins with an even smaller dimension. Noise is initially sampled as \(1 \times S \times 1\) and then concatenated \(B\) times. Each column of size \(1 \times S \times 1\) is sampled from a Gaussian.

%This assortment of noise sampling techniques allows adaptability, enhancing image quality and overall model efficiency. These strategies are pivotal for tuning noise components to meet diverse requirements, ultimately elevating the quality of generated images and contributing to the GAN architecture's effectiveness. 
We show that introducing noise improves model adaptability, enhancing image quality and overall model efficiency.
One point to note is that the ICN and PCN vectors, when concatenated may not represent a Gaussian. Our framework can be extended to incorporate other noise methods as well.

\begin{figure*}[h!]
    \centering
    \includegraphics[width=1.8\columnwidth]{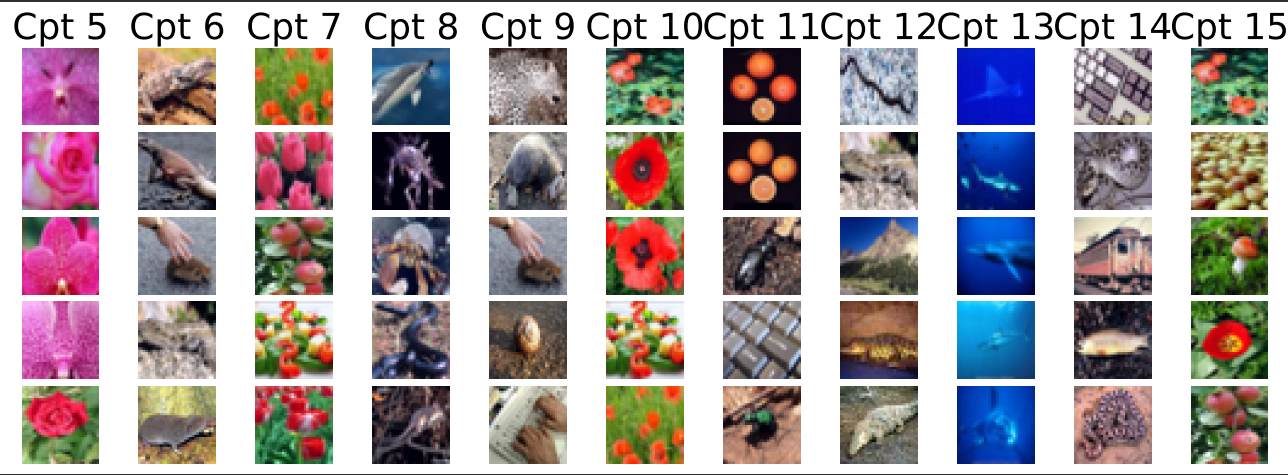}
    \caption{Top 5 images for CIFAR100 that activate the learnt concepts (10 concepts from a subset of 100) using cGAN (VGG 19) DAN  (B=32, S=10). Eg: Cpt 5 corresponds to color pink, Cpt 13 corresponds to object in ocean.}
    \label{fig:cgan_11_cifar100}
\end{figure*}

\subsection{Comparative Analysis}\label{sec:our_results}
% The aim of our experiments is to identify optimal models within each GAN category and various VGG network variations and perform a comparative study over them.
% The robustness of the models is tested by training them five times with distinct seeds and then averaging the accuracies for a final assessment. We use \textit{Accuracy} as our main metric to compare the models. 

Our objective with the experiments was to first identify optimal models within each GAN category and various VGG network variations, and subsequently conduct a comparative study with the baselines. Note that, the backbone architecture remains the same as the baselines, i.e ResNet. To assess the robustness of the models, we employ a training process repeated five times with distinct random seeds. The resulting accuracies are averaged to yield a final assessment.
%Our primary metric for model comparison is \textit{Accuracy}.
We chose a batch size of 32 for processing images, following \cite{Sarkar2021AFF}, to keep the results consistent and comparable. Additionally, the noise length is set to 10, mirroring the size of concepts in the CIFAR-10 dataset, in order to ensure that the noise component makes a substantial contribution to the learned concepts.
%This decision is motivated by the desire to ensure that the noise component makes a substantial contribution to the learned concepts.

% We have opted to use a batch size of 32 for the images as the results from this configuration was also used before in \cite{Sarkar2021AFF}. For the noise length, we wanted the noise to show some significant contribution to the concepts, hence, we chose a size of 10 which is same as the size of concepts for CIFAR-10 data.

\noindent \textit{Label conditioning impact:} \\ Vanilla GAN \cite{GAN} and cGAN \cite{CGAN} are chosen to investigate the impact of label conditioning on our framework. Vanilla GAN generates images from random noise without specific constraints, lacking precise control over image generation. While cGAN leverages labels as an additional input parameter to control the generation of the image. For example, if the network is trained on pictures of different animals, one cannot specify which animal the generator should create.
Some slight modifications are made to the network to ensure its compatibility with both Vanilla GAN and cGAN, ensuring that our framework is adaptable to both types of GANs.

%Vanilla GAN generates images from random noise without specific constraints, while cGAN enables precise generation based on labels. For example, if the network is trained on pictures of different animals, one cannot specify which animal the generator should create.
\textbf{Vanilla GAN:}\label{sec:van_gan} 
%In the context of Vanilla GAN, we will differentiate between different discriminator architectures and choose the best model out each of the architectures with respect to the noise methods.
As shown in Table \ref{tab:all_model}, in the case of \textbf{CIFAR10}, for VGG 8, we observe that ICN  gives the best accuracy of \textbf{91.57\%}. While in the case of VGG 11, PCN  gives better results, with an accuracy of \textbf{91.66\%}. 
As shown in Table \ref{tab:cifar100_compare}, on \textbf{CIFAR100}, for VGG 11 we observe that DAN  has the best accuracy of \textbf{65.49\%}. DAN  also gives the best accuracy of \textbf{60.71\%} for VGG 19. 
%While in the case of VGG 19, DAN (Method 1) has a better accuracy of as compared to other methods.

\textbf{cGAN:}\label{sec:cond_gan} 
%In the context of cGAN, we will differentiate between different discriminator architectures and choose the best model out each of the architectures with respect to the noise methods. 
Here, in addition to a noise size of 10, we also experiment with a noise size of 5 to examine its effect on the framework.
As shown in Table \ref{tab:all_model}, for \textbf{CIFAR10}, using VGG 8, PCN gives the best accuracy of \textbf{91.60\%}. In the case of VGG 11, for a noise size of 10, we get better results using DAN - with an accuracy of \textbf{91.58\%}; whereas for a noise size of 5, 
we get better results using PCN - with an accuracy of \textbf{91.47\%}. Finally, in the case of VGG 19, DAN gives the best results when using a noise size of 5.
As shown in Table \ref{tab:cifar100_compare}, for \textbf{CIFAR100}, we consistently observe that DAN gives the best results. In the case of VGG 11, using DAN we get the best accuracy of \textbf{65.37\%}. With VGG 19, using DAN we get the best accuracy of \textbf{64.42\%}.\\

\noindent So, we see that different noise methods work well on different models, based on the model
parameters such as depth, nonlinearity, connections etc, with the best methods being chosen for comparison with the baselines. Table \ref{tab:cifar100_all} and Table \ref{tab:res} show our frameworks results on CIFAR100 and CIFAR10 respectively for experiments using multiple architectures and parameters. We can see that our framework performs better than the baselines. We observe from the table, that the best performing model in terms of accuracy is \textbf{cGAN} with \textbf{VGG 19} as the discriminator, and a noise size 5, giving an accuracy of \textbf{91.82\%}. For CIFAR100, it is observed that the best performing model in terms of accuracy is \textbf{Vanilla GAN} with \textbf{VGG 11} as the discriminator using DAN giving an accuracy of \textbf{65.49\%}, although the configuration with the best auxiliary accuracy is given by a \textbf{cGAN} with VGG 11 using DAN giving \textbf{45.36\%}. \\

\noindent \textit{Concept Visualization:}\\ We visualize the top 5 images where a particular concept $\lambda_i$ had the highest score as compared to the other concepts. So, the images \textit{visualize} the captured concepts, or \textit{activate} a particular concept. We also show that concepts are captured across classes. Using this method, we show the concepts captured by our model using CIFAR10 in Fig. \ref{fig:cgan11} and our model using CIFAR100 in Fig. \ref{fig:cgan_11_cifar100}. %We also show concept visualizations for some other good models - Figure \ref{fig:cgan11} for cGAN with VGG 11 and Figure \ref{fig:vgan8} for Vanilla GAN with VGG 8. 

\subsection{Implementation Details}

%Our primary goal was to explore novel ways of incorporating GANs within an Ante-Hoc context. The conventional decoder is replaced with a GAN-like structure, to provide a larger search space for the generator. This enables the architecture to incorporate more flexibility.

%Vanilla GAN and cGAN are chosen to investigate the impact of label conditioning on our framework. Vanilla GAN lacks control over image generation, while cGAN leverages labels as an additional input parameter to control the generation of the image. Some slight modifications are made to the network to ensure its compatibility with both Vanilla GAN \cite{GAN} and cGAN \cite{CGAN}. An important feature of the architecture is that it is adaptable to both types of GAN, making the model more robust and versatile. Vanilla GAN generates images from random noise without specific constraints, while cGAN enables precise generation based on labels. For example, if the network is trained on pictures of different animals, one cannot specify which animal the generator should create.

The generator architecture comprises multiple deconvolution layers, generating images using learned concepts and noise, and incorporating labels in the case of cGAN. The discriminator architecture has a VGG network backbone with a few additional layers to re-purpose it into a binary classifier. We have performed tests with different VGG \cite{VGG} architectures such as VGG 8, VGG 11 and VGG 19. 
%We have studied the effects of label conditioning, and it has provided insights into integrating diverse information within our framework. 

\subsection{Observations}\label{sec:observations}
The preceding sections discussed how various VGG models can influence the performance of our framework. Our observations underline that GAN-based conditioning, cGAN, introduces notable improvements. 
%This results from the conditioning's inherent capacity to control image generation, thereby enhancing the model's predictive outcomes.
A general trend observed indicates that VGG model depth correlates with improved performance, particularly in terms of accuracy. 
%Our study reveals that among the methods, DAN coupled with cGAN and VGG 19 yields the most favourable results with an accuracy of \textbf{91.82\%}. 
%Additionally, in many instances, the alternative methods perform similar to \cite{Sarkar2021AFF} when comparing the Accuracy.
We also see a correlation between dataset scale and auxiliary accuracy, from the fact that our method consistently gives better auxiliary accuracy compared to the baselines on CIFAR100.

Another significant observation is that the increasing complexity of the model due to the GAN integration and noise sampling methodologies (as detailed in Section \ref{sec:data_noise}); increase training time by 1.4 times that of \cite{Sarkar2021AFF}. Despite this, the training efficiency remains considerably superior to that of SENN \cite{SENN}.

\section{Conclusion and Future Work}\label{sec:conclusion}
In conclusion, this work presents a method for incorporating a Generative Adversarial Network (GAN) into an ante-hoc explainability framework. The design replaces a conventional decoder network with a GAN and fine-tunes the framework. The exploration of noise sampling methods, specifically the implementation of DAN, demonstrate superior performance, proving the effectiveness of a GAN in aiding the process of encoding concepts. 
We have observed results that signify an improved overall accuracy and auxiliary accuracy, highlighting the potential of our architecture for robust image classification and effective concept learning. 
%We also observe a positive correlation between the size of the VGG network in the discriminator and the accuracy achieved could guide future architectural choices.

Although, we have made some improvements on enhancing the explainability of deep neural networks without losing out on classification performance, the work presented is a small step towards a much more robust and human interpretable model.
%In the past, such Deep Neural Network models did not have the capability to explain their predictions, whereas now the field has seen a lot of advancements. 
In the future, we plan on exploring the possibilities of more advanced and complex architectures that could involve using a much deeper classification models such as ResNet, EfficientNet, Mask RCNN. We also plan on making use of the capabilities of some state-of-the-art architectures such as Vision Transformers in conjunction with our framework.

\section{Acknowledgements}
We would like to thank the authors of \cite{Sarkar2021AFF} for their guidance and insightful discussions. We would also like to thank the anonymous reviewers for their valuable feedback.

\bibliography{aaai24}

\end{document}